\documentclass[runningheads]{llncs}
\usepackage[T1]{fontenc}
\usepackage{graphicx}

\usepackage{booktabs}
\usepackage{tabularx}
\usepackage[skins,breakable]{tcolorbox}
\usepackage{adjustbox}
\usepackage[acronym]{glossaries}
\usepackage{multirow}
\usepackage{flushend}
\usepackage{marvosym}
\usepackage{orcidlink}

\begin{document}

\title{Zoom In Disparities in Healthcare LLM Q\&A}





\author{Ipek Baris Schlicht\inst{1}\textsuperscript{\Letter}\orcidlink{0000-0002-5037-2203} \and
Burcu Sayin \inst{2}\orcidlink{0000-0001-6804-127X} \and
Zhixue Zhao \inst{3}\orcidlink{0000-0002-3060-269X} \and \\
Frederik M. Labonté\inst{4,5} \and
Cesare Barbera\inst{6}\orcidlink{0009-0004-5568-6166} \and \\
Marco Viviani\inst{7}\orcidlink{0000-0002-2274-9050} \and
Paolo Rosso\inst{1,8}\orcidlink{0000-0002-8922-1242} \and
Lucie Flek\inst{4,5}\orcidlink{0000-0002-5995-8454}} 
%

\institute{Universitat Politècnica de València, Valencia, Spain 
\and University of Trento, Trento, Italy \and
University of Sheffield, Sheffield, UK \\\and
Bonn-Aachen International Center for Information Technology, University of Bonn, Bonn, Germany \\
\and
Lamarr Institute for Machine Learning and Artificial Intelligence, Germany, Dortmund, Germany \and
University of Pisa, Pisa, Italy \\
\and
University of Milano-Bicocca, Milano, Italy \\
\and
ValgrAI Valencian Graduate School and Research Network of Artificial Intelligence, Valencia, Spain \\
\textsuperscript{\Letter}\email{ibarsch@doctor.upv.es}
}

\authorrunning{Schlicht et al.}
%


\maketitle

\newif\ifproofread
\newcommand{\changemarker}[1]{%
\ifproofread
\textcolor{red}{#1}%
\else
#1%
\fi
}

\newcommand{\revtable}{%
  \ifproofread
    \color{red}%
  \fi
}

\proofreadfalse

\begin{abstract}
This paper systematically examines cross-lingual disparities in pre-training source and factuality alignment in Large Language Model (LLM) answers for multilingual healthcare Q\&A across English, German, Turkish, Chinese (Mandarin), and Italian.
To support this analysis, we $(i)$ constructed MultiWikiHealthCare, a multilingual dataset derived from Wikipedia; $(ii)$ used it to examine cross-lingual differences in healthcare-related coverage; $(iii)$ evaluated the alignment between LLM-generated responses and these reference sources; and $(iv)$ conducted a case study on factual alignment through the use of contextual information and Retrieval-Augmented Generation (RAG).
Our findings reveal substantial cross-lingual disparities in both Wikipedia coverage and LLM factual alignment. \changemarker{Based on our Wikipedia analysis, the lowest alignment is observed between English and Chinese Wikipedia pages. }Across models, responses align more with English Wikipedia, even when the prompts are non-English. We further show that providing contextual excerpts from non-English Wikipedia at inference time effectively shifts factual alignment toward target knowledge. 
\end{abstract}

\section{Introduction}

\textit{Large Large Language Models} (LLMs) are increasingly deployed across healthcare applications, and people seeking health information routinely turn to LLM-based systems for advice and guidance~\cite{yagnik2024medlm,yu2024enhancing}. Since LLMs are trained primarily on large-scale online data, their responses are shaped by the availability and quality of online health information~\cite{nigatu2025into}. However, this information varies markedly across languages, reflecting disparities in health communication, infrastructure, and cultural norms~\cite{tierney2025health,jbp:/content/journals/10.1075/dt.25015.yan}. 

Many healthcare benchmarks probe both LLM hallucination and disparity analysis~\cite{agarwal2024medhalu,koopman-zuccon-2023-dr,kim2025medical,DBLP:conf/www/ZhuA0R19,samir-etal-2024-locating}, but they are largely English-centric or too coarse-grained to diagnose performance gaps across languages. Moreover, although related, the two concepts operate at different levels of analysis. While hallucination detection focuses on identifying content that is factually incorrect or fabricated \cite{kim2025medical,DBLP:journals/corr/abs-2309-01219}, \textit{disparity analysis} examines how information is differently represented or prioritized across linguistic and contextual boundaries, even if the facts themselves might vary
across different cultural/contextual settings \cite{samir-etal-2024-locating,ranathunga-de-silva-2022-languages}.

\begin{figure*}[!t]
\tiny
    \centering
\includegraphics[width=\textwidth]{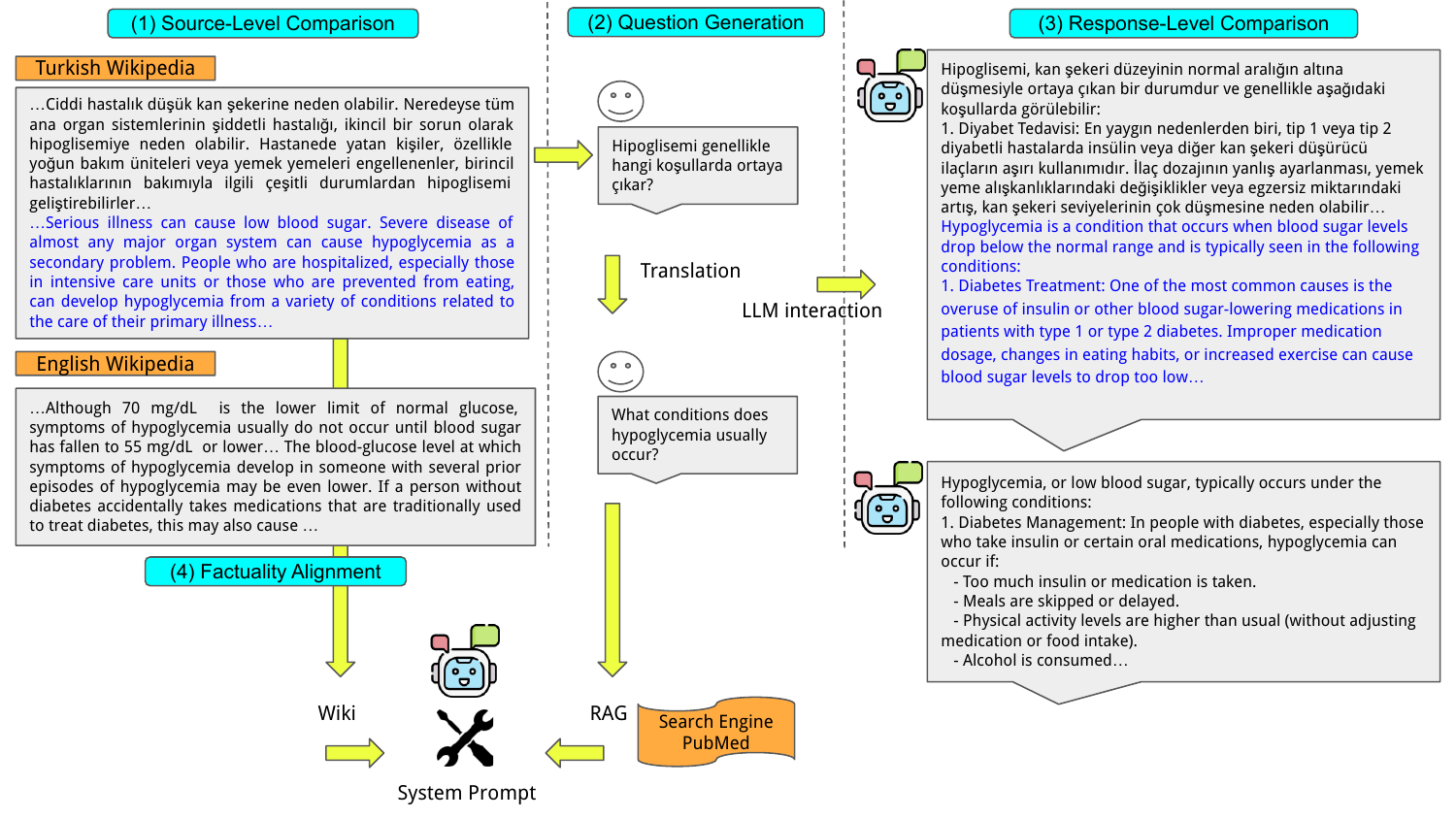}
    \caption{The framework for examining source- and response-level disparity and factuality alignment: (1) Comparison of Wiki pages (TR-EN) in terms of factuality and structure; (2) Question generation based on facts from the Wiki pages; (3) Response factuality evaluation; (4) Contextual alignment using Wiki pages and RAG. The EN translations are given in \textcolor{blue}{blue}.}
    \label{fig:teaser}
\end{figure*}

Recent studies have applied disparity analysis to Wikipedia entries on people and cuisines~\cite{samir-etal-2024-locating,wang2025wikigap}, and others have investigated disparities in LLM outputs within the medical domain~\cite{gupta2025found,restrepo2025multi}. However, none of these works explicitly link disparities between the source (e.g., Wikipedia) and LLM-generated responses. This paper introduces a holistic framework to assess how LLM-generated health answers align with factuality and language-specific information across languages. As illustrated in Figure~\ref{fig:teaser}, $(i)$ we begin by comparing healthcare-related Wikipedia pages, which is a pretraining corpus for many LLMs~\cite{singhal_large_2023},  across languages to characterize similarities and discrepancies in coverage, phrasing, and citation patterns. From this analysis, $(ii)$ we construct aligned cross-lingual fact sets and use them to generate questions posed to several multilingual LLMs, i.e., Llama3.3-70B~\cite{dubey2024llama}, Qwen3-Next-80B-A3B-Instruct~\cite{yang2025qwen3}, and Aya~\cite{dang2024aya}. Subsequently, $(iii)$ we 
evaluate the responses for quality and alignment with both English Wikipedia and the corresponding target-language pages. Finally, $(iv)$ we present a preliminary case study testing whether providing non-English contextual excerpts at inference time shifts factual alignment toward locally relevant sources, through the use of contextual information and \textit{Retrieval-Augmented Generation} (RAG).

The main contributions of this work can be summarized as follows: $(i)$ We constructed \textbf{MultiWikiHealthCare}, a multilingual dataset curated from trending health-care topics and Wikipedia, covering English (EN), German (DE), Italian (IT), Turkish (TR), and Chinese (ZH). It enables systematic comparisons of disparities at both the source level (Wikipedia) and the response level (LLM outputs). $(ii)$ Analysis of Wikipedia reveals substantial cross-lingual disparities relative to EN, with ZH showing the lowest alignment and fewest extracted facts; DE more often cites regional sources, while others rely on international sources. $(iii)$ LLM outputs exhibit pronounced EN-centric alignment: responses track English Wikipedia more closely than same-topic pages in other languages, with further drops when queries are posed in EN. It might be problematic for culturally specific knowledge with differing practices and guidelines. $(iv)$ A context-augmented prompting case study shows LLMs can shift alignment toward non-EN sources at inference time, highlighting the value of incorporating target-language knowledge. Finally, our dataset and source code \changemarker{with system prompts are available on GitHub~\footnote{\url{https://github.com/isspek/nldb2026-multiwikihealthcare}}.}

\section{Related Work}\label{sec:related_work}

Research on factual reliability, hallucination detection, and cross-lingual consistency in LLMs provides important insights into the origins of disparity in model behavior across languages and domains. 
Prior work shows that prompt design strongly influences factuality in medical Q\&A~\cite{koopman-zuccon-2023-dr,kim2025medical,sayin-etal-2024-llms}.However, while LLMs rarely contradict medical facts, they often fail to challenge incorrect ones~\cite{kaur-etal-2024-evaluating}. Extending to multilingual settings, accuracy, consistency, and verifiability vary substantially across languages~\cite{jin2024better}.

MedHalu and the MedHaluDetect framework target fine-grained hallucination detection in medical responses~\cite{agarwal2024medhalu}. Built from English healthcare questions spanning HealthQA, LiveQA, and MedicationQA~\cite{DBLP:conf/www/ZhuA0R19,DBLP:conf/trec/AbachaAPD17,DBLP:conf/medinfo/AbachaMSGSD19}, the dataset incorporates hallucination taxonomies~\cite{DBLP:journals/corr/abs-2309-01219} via synthetic perturbations generated with GPT-3.5~\cite{achiam2023gpt}. Evaluations show LLMs underperform both experts and lay users in hallucination detection.

To analyze cross-lingual information gaps in Wikipedia, Samir et al.~\cite{samir-etal-2024-locating} proposed InfoGap, which decomposes and aligns facts across languages using GPT-4, applied to biographies in the LGBTBio Corpus~\cite{park2021multilingual}. In healthcare, multilingual disparities have been examined in mental-health responses via translation-based evaluation~\cite{gupta2025found} and in ophthalmological Q\&A benchmarks assessing multiple-choice accuracy~\cite{restrepo2025multi}, which also introduced CLARA for inference-time debiasing using RAG and self-verification. Another research reports cross-lingual variation in detail, numerical consistency, contradictions, irrelevance, and citation practices~\cite{schlicht2025llms}. \changemarker{Unlike prior work~\cite{gupta2025found,restrepo2025multi,schlicht2025llms}}, we study the relationship between factual knowledge potentially acquired during pretraining and the factual alignment of LLM-generated responses across languages through a dedicated benchmark. \changemarker{In comparison with CLARA, our dataset spans multiple healthcare topics.} Furthermore, we extend this analysis to culturally diverse languages. 

\section{Multilingual Wiki Health Care}

\begin{figure}[!t]
\centering
\scriptsize
\includegraphics[width=0.6\textwidth]{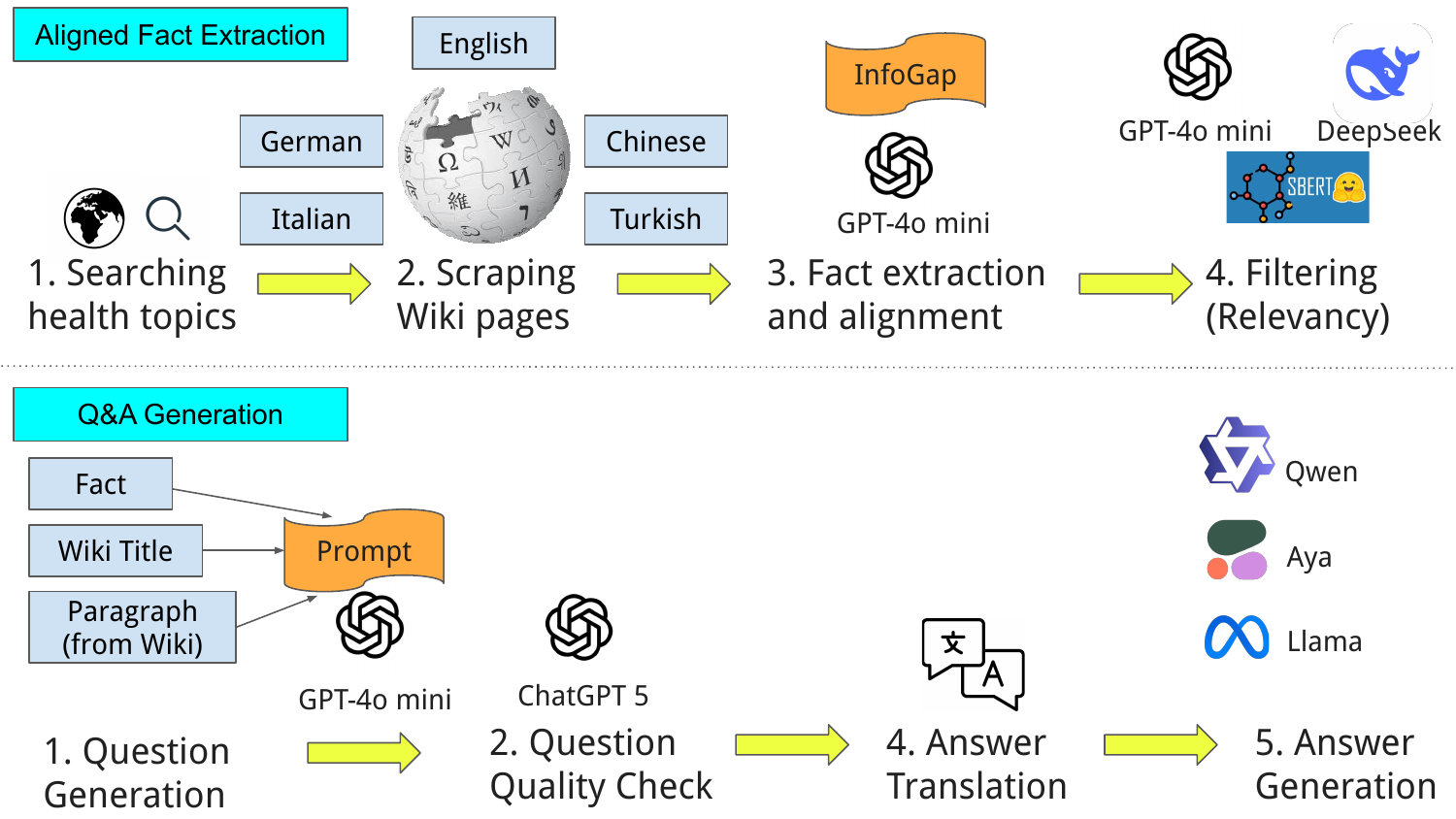}
\caption{The pipeline for Q\&A construction.}
    \label{fig:ds_pipeline}
\end{figure}

Existing benchmark datasets largely focus on monolingual evaluation, typically for general-purpose or other tasks such as hallucination detection, lacking the specificity needed to assess how pretraining sources affect multilingual answer quality in specialized domains such as healthcare. To address this gap, we introduce \textbf{ MultiWikiHealthCare}, a multilingual, health-focused Q\&A dataset. MultiWikiHealthCare is derived from Wikipedia, a common pretraining source for LLMs. Figure~\ref{fig:ds_pipeline} presents the pipeline for constructing the dataset. The main stages of the pipeline are outlined in detail in the remainder of this section.

\subsection{Construction of Aligned facts}
\noindent
\textbf{Healthcare Topics}
To construct MultiWikiHealthCare, we first used Google Trends in 2025 to identify high-salience health topics.\footnote{\url{https://trends.google.com/trends/explore}} To ensure broader coverage, we manually expanded it with additional health-related topics from a Wikipedia list of \textit{controversial issues} in science, biology, health,\footnote{Wikipedia:List\_of\_controversial\_issues\#Science,\_biology,\_and\_health} and as well as topics from a related survey~\cite{schlicht_automatic_2024}. These topics formed the basis of the content collection. The topics are \textit{Allergy}, \textit{Cancer}, \textit{Cardiovascular Disease}, \textit{Cold}, \textit{Covid}, \textit{Depression}, \textit{Diabetes}, \textit{Diet}, \textit{Ebola}, \textit{Flu}, \textit{Influenza}, \textit{Nutrition}, \textit{Obesity}, \textit{Pain}, \textit{Smoking}, \textit{Vaccination}, \textit{Weight Loss}. We used Google Trends to identify their sub-topics from related entities trending between 2004 and 2025 across global and country-level search data (U.S., U.K., Turkey, Germany, Italy, and China). Across all languages, we identified 1,193 unique entities. Symptoms, causes, and diseases are common entities across languages, while some entities are not specific to the healthcare domain.


\noindent
\textbf{Scraping Wikipedia Pages}
We used Llama 3.3-70B and served through the vLLM inference framework~\footnote{\url{https://docs.vllm.ai/en/latest/}}~\cite{kwon2023efficient}, to $(i)$ filter out entities not related to healthcare and $(ii)$ link the remaining entities to their corresponding Wikipedia pages and then removed the duplicate pages. Wikipedia page titles often differ across languages, especially when scripts differ or the title is a common noun rather than a proper name (e.g., EN/IT “Nausea” vs. DE “Übelkeit”). herefore, we retrieved \textit{interlanguage titles} using the open-source Wikipedia API.\footnote{\url{https://github.com/martin-majlis/Wikipedia-API}} Pages available only in EN were excluded as they are unsuitable for comparative analysis, yielding 815 titles present in EN and at least one additional language. A subsequent manual review showed that some pages referred to `films', `people', or `doctoral degree'. Hence, we removed Wikipedia pages referring to such category metadata. 
\\
\noindent
\textbf{Fact Extraction and Alignment}\label{sect:fact_alignment}
\begin{table}[!t]
    \caption{Comparision of human \changemarker{annotations}, InfoGap predictions and random guess.}
\small
    \centering
    \begin{tabular}{lcc}
    \toprule
    \textbf{Language Pair} & \textbf{F1-Macro} & \textbf{Random}\\
    \midrule
         en $\leftrightarrow$ TR & 0.841 & 0.574 \\
         en $\leftrightarrow$ ZH & 0.724 & 0.535 \\
         en $\leftrightarrow$ DE & 0.684 & 0.479 \\
         en $\leftrightarrow$ IT & 0.638 & 0.538 \\
    \bottomrule
    \end{tabular}
    \label{tab:lang_pair}
\end{table}

From the Wikipedia articles, we extracted \textit{facts} together with their supporting paragraphs as evidence. We categorized these facts into two groups: (1) \textit{Cross-lingual overlapping facts}, where the same factual content appears in both EN and non-EN version (2) \textit{Language-specific facts}, which are unique to the non-EN article without an EN counterpart. 
The first group is used to construct the Q\&A dataset and the second is part of the Wikipedia analysis discussed in Section~\ref{sec:wiki_method}.

Fact extraction and alignment were performed with InfoGap~\cite{samir-etal-2024-locating,wang2025wikigap}, the state-of-the-art framework for cross-lingual fact extraction and alignment on Wikipedia (see Section \ref{sec:related_work})
. The framework combines OpenAI LLMs with a hubness-based correction technique to improve cross-language alignment accuracy. Because a single Wikipedia page can contain hundreds of atomic facts, running OpenAI models at a corpus scale is costly~\cite{samir-etal-2024-locating}. While the latest InfoGap~\cite{wang2025wikigap} employs GPT-4o~\cite{achiam2023gpt}, we adopted GPT-4o-mini~\cite{hurst2024gpt} as the backbone model, which is approximately 10\% less expensive than GPT-4o. To assess reliability on our corpus, we sampled 50 facts per language following the InfoGap evaluation protocol. Annotations were conducted by volunteer native speakers using the official InfoGap guidelines. Our results (see Table~\ref{tab:lang_pair}) are comparable to those reported by \cite{samir-etal-2024-locating}, which outperform a random prediction.\footnote{In \cite{samir-etal-2024-locating} random guessing outperformed Natural Language Inference Transformers.}

\begin{table}[!t]
    \caption{Transformers (TF) against GPT-4o-mini (F1-macro)}
\small
    \centering
    \begin{tabular}{cccc}
\toprule
        \textbf{Language} & \textbf{LLM} & \textbf{Mono TF} & \textbf{Cross TF} \\
    \midrule
        EN & 82.69 & 86.49 & \textbf{88.44} \\
        TR & 77.89 & 83.46 & \textbf{84.20} \\
        ZH & 68.47 & \textbf{77.02} & 72.11 \\
        DE &  76.67 & 70.88 & \textbf{81.32} \\
        IT & 71.96 & 72.00 & \textbf{83.33} \\
\bottomrule
    \end{tabular}
    \label{appx:rel_result}
\end{table}
\noindent
\textbf{Selecting Relevant Facts} \label{subsec:relevant_check} After cross-lingual extraction and alignment, we retained only bidirectionally matched facts: the intersection of the two directions for each language pair (e.g., EN $\leftrightarrow$ TR). Although the facts are atomic, InfoGap returns the aligned source sentences in both languages. We then located the paragraphs containing these sentences and matched each fact with its corresponding bilingual evidence.

We observed that not all Wikipedia-derived facts are relevant to health information seekers (e.g., some facts are historical or highly technical). For instance, statements such as ``Vitamin C exhibits low/acute toxicity'' are relevant, whereas others like ``If a dog was sick, they would get better food'' are not.. To remove irrelevant content, we implemented a relevance filter. As the corpus is large, running GPT-4o-mini over all facts would be time-consuming and costly. To lower inference cost while preserving quality, we distilled GPT-4o-mini’s judgments (and DeepSeek-R1~\cite{guo2025deepseek}’s for ZH) into a smaller model.

To calibrate task understanding and design the relevancy prompt, we performed annotations. \changemarker{First five authors} labeled \changemarker{initially} at least 25 EN samples, with two annotators per sample. We held regular discussions to refine the annotation guidelines. Inter-annotator agreement, measured by Krippendorff’s alpha~\cite{ford2004content} on a subset of 52 samples, was $\alpha=0.72$. Each \changemarker{of us} then labeled 50 samples in their respective language. The annotations form the test set for comparing the LLMs with transformer-based models across languages. The finalized EN prompt was translated into other languages by native speakers.

We annotated 2,000 TR and EN, 4,000 DE, 3,000 IT, and 3,000 ZH samples with LLMs. GPT-4o-mini was used for all except ZH, which was labeled with DeepSeek-R1 due to poor alignment. Each dataset was randomly split into 70\% training and 30\% development sets. We fine-tuned language specific transformers on their monolingual dataset (Roberta-base~\cite{liu2019roberta} for EN, BertTurk~\footnote{\url{https://huggingface.co/dbmdz/bert-base-turkish-cased}} for TR, German BERT~\cite{devlin-etal-2019-bert} for DE) and  Chinese BERT~\cite{devlin-etal-2019-bert} for the ZH dataset. Additionally, each non-EN dataset was combined with the EN samples, and XLM-RoBERTa was then fine-tuned~\cite{xlmroberta} on the respective combined dataset (e.g., the TR and EN datasets were combined to predict the TR samples using XLM-RoBERTa). All models were fine-tuned with a learning rate of 2e-5, for 3 epochs, and a batch size of 16. The models with the best F1-macro scores (see Table~\ref{appx:rel_result}) were selected. 
Except for ZH, cross-lingual models performed best and were used as relevance classifiers. Finally, we labeled the corpora with the fine-tuned transformers~\cite{wolf2020transformers} and discarded unrelated samples. 

\subsection{Question and Answer Generation}

\noindent
\textbf{Question Generation} We generated synthetic health-related questions using a prompt that instructed GPT-4o-mini to act as a health information seeker. The prompt took as input a Wiki page name, a fact, and its paragraph. From the dataset mentioned in Section \ref{subsec:relevant_check}, we sampled 1,100 samples per language to generate.

Formally, let each data instance be represented as $d = (f_x, f_{en}, p_x, p_{en})$, where $f_x$ is a fact in Language $X$, $f_{en} = \{ f_{en}^1, \dots, f_{en}^n \}$ ($n \geq 1$) is its aligned EN fact(s), $p_x$ is the paragraph in Language $X$ containing $f_x$, and $p_{en}$ is the aligned EN paragraph(s). Given $(f_x, p_x, p_{en})$, we used GPT-4o-mini to generate a question $q_x$ in Language $X$ and then translated it to EN with Google Translate to get the question pair.

We evaluated the quality of the generated questions based on the LLM-as-a-Judge technique \cite{gu2025surveyllmasajudge} using ChatGPT-5 (Extended Thinking)~\cite{gpt5_2025}. This prompt instructed ChatGPT-5 to execute a deterministic Python-based evaluation pipeline within its data-analysis sandbox \cite{gpt5_2025}. The model applied four binary criteria to each question: (1) relevance to both the input fact and the source paragraph, (2) answerability based solely on the paragraph, (3) alignment with natural health-seeker intent, and (4) clarity of expression.

To assess alignment with human judgments, we sampled 20 items per language and compared ChatGPT-5’s decisions to native-speaker annotations. The agreement ranged from 44\% to 76\% (highest for TR, lowest for DE). ChatGPT-5 consistently accepted fewer questions than human annotators, indicating a more conservative criterion. Based on this preliminary evidence, we used ChatGPT-5 as a pre-filter for question quality.
\\
\noindent
\textbf{Answer Generation} To generate answers, we use multilingual, open-weight LLMs from distinct organizations to mitigate model-family bias in our evaluation.The models are Llama 3.3–70B (Meta; an upgraded release of Llama 3~\cite{grattafiori2024llama}), Qwen (Qwen3-Next-80B-A3B-Instruct)~\cite{yang2025qwen3} from Alibaba Cloud and Aya (Expanse-32b) from Cohere~\cite{dang2024aya}. We discarded DeepSeek-R1 from the analysis due to its size and cost. We run Aya locally on a GPU, while the other models are accessed via APIs through Hugging Face Inference.\footnote{\url{https://huggingface.co/docs/huggingface\_hub/en/package\_reference/inference\_client}} For all LLMs, we set the temperature at 1 and the maximum token length to 4,096. The final dataset comprises 854 TR, 997 DE, 502 IT, and 548 ZH samples. We manually reviewed a subset of Q\&A pairs: most answers are relevant, with only a few misalignments due to broader questions than the available Wikipedia evidence.


\section{Experiments and Results}
Main research question in this paper is how disparities across languages in pre-training data contribute to inconsistencies in LLM answers for multilingual healthcare Q\&A.
We first characterize healthcare-related Wikipedia pages, we then evaluate each model’s answers for cross-lingual factual alignment against evidence extracted from the corresponding pages.

\subsection{Comparison of Wikipedia Pages}\label{sec:wiki_method}

We examine the amount of information presented in Wikipedia articles across languages. To this extent, we compare number of sections, paragraphs, facts and external links that the articles cited.

As articles are typically organized into sections that aid navigation and reflect the logical structure of the content, we use the number of sections as a simple, language-agnostic proxy. Accordingly, we measure and compare section counts across languages.
We use Beautiful Soup to parse each article’s HTML.\footnote{\url{https://beautiful-soup-4.readthedocs.io/en/latest/}} Next, we count the paragraphs and facts obtained through InfoGap, as described in Section~\ref{sect:fact_alignment}, with the results summarized in Table~\ref{stats:ds}. Lastly, we analyze the references cited by Wikipedia editors to examine how reference preferences vary across languages. References serve as important indicators of information diversity and reliability. We restrict our analysis to entries with articles in all target languages. We begin by comparing the number of references per article across languages and then compare the sources cited by each edition. Additionally, we extract the sources of the external links by using \texttt{tldextract}.\footnote{\url{https://github.com/john-kurkowski/tldextract}}

\begin{table}[!ht]
\caption{Statistics of paragraphs, facts and aligned facts on the Wiki pages of MultiWikiHealthCare. NA is Not Applicable. Many EN facts don't exist in other language editions. ZH Wiki has the lowest aligned facts with the EN wiki pages.}
\small
\centering
\adjustbox{max width=\columnwidth}{
    \begin{tabular}{cccccc}
    \toprule
         &  & & \multicolumn{2}{c}{\textbf{Aligned facts (\%)}} \\
         \textbf{Language} & \textbf{Paragraphs} & \textbf{Facts} & \textbf{EN $\rightarrow$ Target} & \textbf{Target $\rightarrow$ EN} \\
    \midrule
         EN & 26,752 & 205,468 & NA & NA \\ 
         \midrule
         TR & 8,554 & 47,711 & 33.79\% & 91.82\% \\
         ZH & 728 & 6,155 & 6.98\% & 59.46\% \\
         DE & 18,284 & 118,268 & 23.36\% & 98.44\% \\
         IT & 14,846  & 96,342 & 26.73\% & 54.23\% \\
    \bottomrule
    \end{tabular}}
    \label{stats:ds}
\end{table}

\begin{table}[ht]
\centering
\revtable
\scriptsize
    \caption{\changemarker{The English (EN) Wiki pages contain statistically more information than their other editions (Paired t-test $p$ value)}.}
    \begin{tabular}{lcccc}
    \toprule
         & \textbf{Sections} & \textbf{Paragraphs} & \textbf{Facts} & \textbf{Links} \\
    \midrule
       TR-EN &  \(1.51\times10^{-96}\) &  \(2.88\times10^{-87}\) & \(1.72\times10^{-106}\) & \(6.24\times10^{-68}\) \\
       DE-EN &  \(2.48\times10^{-22}\) & \(3.39\times10^{-28}\) & \(1.99\times10^{-48}\) & \(7.14\times10^{-79}\) \\
       ZH-EN &  \(1.44\times10^{-77}\) & \(1.39\times10^{-141}\) & \(1.59\times10^{-133}\) & \(6.55\times10^{-60}\) \\
       IT-EN & \(1.35\times10^{-50}\) & \(1.80\times10^{-50}\) & \(9.11\times10^{-70}\) & \(4.75\times10^{-74}\) \\
    \bottomrule
    \end{tabular}
    \label{tab:wiki_analysis}
\end{table}

Amount of information (sections, paragraphs, facts and links) in EN Wiki pages, according to paired t-test result~\cite{ross2017paired}, is statistically more than their pages in other languages \changemarker{(in Table~\ref{tab:wiki_analysis})}. EN edition contains the most paragraphs, followed by DE and IT. It also yields substantially more extracted facts than other editions, and many EN facts lack counterparts in other languages. Among all editions, ZH Wikipedia has the fewest facts aligned with EN. A plausible explanation for this discrepancy is that the EN edition draws on a broader, more globally distributed editor base than other language editions.

Across languages, domains associated with the National Institutes of Health and the Centers for Disease Control and Prevention are common. Other high-frequency domains are news outlets, scholarly journals, and file archives. Distinctively, the DE edition also cites national sources (e.g. rki.de, aerzteblatt.de). In summary, citation coverage and practices vary substantially across language editions, reflecting differences in local information ecosystems and editorial norms.

\subsection{Analyzing Answers}\label{exp:answer_analyze}
\begin{table}[!t]
\small
\caption{Factual alignment between answers and Wiki excerpts in the source-language ({non-EN}) and English ({EN}), measured with AlignScore. Non-EN content was translated into EN. Answers generally align more with EN pages, while EN questions show lower similarity to source-language references.}
\centering
\adjustbox{max width=\columnwidth}{
\begin{tabular}{ccccccc} 
\toprule
\textbf{Query} & \multicolumn{2}{c}{\textbf{Llama}} & \multicolumn{2}{c}{\textbf{Aya}} & \multicolumn{2}{c}{\textbf{Qwen}} \\
\midrule
 & non-EN & EN &  non-EN & EN  & non-EN & EN \\
 TR    &  27.44 & 30.89 & 17.78 & 22.62 & 14.45 & 18.31 \\
 EN & 16.96 & 21.64 & 15.03 & 20.02 & 14.04 & 18.28  \\
                         \midrule
 DE    & 18.13 & 18.56 & 16.26 & 17.92 & 13.64 & 16.59\\
EN & 16.85 & 18.68 & 14.69 & 17.37 & 13.38 & 17.02  \\
                        \midrule
 ZH    & 22.39 & 28.13 & 20.22 & 26.78 & 17.45 & 22.71 \\
EN & 20.09 & 26.23 & 16.90 & 23.81 & 15.58 & 21.72\\
                        \midrule
 IT    & 20.16 & 23.51 & 16.40 & 20.94 & 14.12 & 18.41 \\
EN & 16.65 & 21.64 & 15.04 & 19.97 & 13.54 & 17.97 \\
\bottomrule
\end{tabular}}
\label{exp:sec2_answer_sim}
\end{table}
Since our analysis tools are primarily designed for EN, we translated all non-EN answers and their supporting evidence into EN prior to scoring. We begin the analysis with the response length. Across non-EN prompts, Qwen consistently produced longer answers, reflecting a tendency toward more elaborated outputs. 
For questions originally written in ZH, most models generated relatively short responses, except for Qwen, which produced notably longer ones. Qwen might be trained on a larger proportion of ZH data compared to the others.

We evaluate answer pairs in EN and other languages against their corresponding Wikipedia evidence using AlignScore~\cite{zha-etal-2023-alignscore}, which computes sentence-level factual consistency by matching each answer sentence to all evidence segments (“context chunks”) with a RoBERTa alignment model~\cite{liu2019roberta}. Because many questions originated in non-EN languages, the aligned EN evidence often comprises multiple passages; when this occurs, we use the passage that yields the highest AlignScore for the given answer.
Answers from conversational LLMs frequently include background and discursive material beyond the minimal facts needed to address a question~\cite{xu-etal-2022-answer}, whereas our evidence snippets are concise and fact-dense. Consequently, absolute AlignScores tend to be modest; we interpret them as a relative proxy for factual alignment rather than a comprehensive quality metric. Therefore, we expect higher scores when a response is factually similar to the source-language reference, and lower scores otherwise.

Table~\ref{exp:sec2_answer_sim} shows that the generated answers are factually close to EN references in most cases. However, when questions are asked in EN, factual alignment decreases for all languages except DE. In particular, factual alignment is higher when evaluated against source-language evidence than against evidence from the EN Wikipedia pages. This finding is consistent with prior work showing that responses to equivalent prompts can diverge across EN and non-EN settings~\cite{jin2024better,schlicht2025llms}.

Lastly, we used the Spearman correlation coefficient~\cite{wissler1905spearman} ($\rho$) to assess the relationship between Wikipedia page quality and LLM answer quality across languages. For this analysis, we additionally measured the relevance of the answers to the questions by using \texttt{ragas}~\cite{es-etal-2024-ragas} with GPT-40-mini. Correlations were generally weak to moderate ($\rho = 0.01-0.34$). For it, tr, and de, the relationships were negligible or weak ($\rho < 0.20$), suggesting that the number of sections, paragraphs, or facts in the corresponding Wikipedia entries had limited predictive power for the factuality, relevance, or length of model outputs. In contrast, the correlations to the factuality scores for ZH were higher, reaching moderate strength for Aya ($\rho \approx 0.34$) and Llama ($\rho \approx 0.30$), and weak-to-moderate for Qwen ($\rho \approx 0.27$). Wikipedia content quality in ZH Wikipedia might have positive effect on factuality of the responses.

\subsection{Case Study: Factuality Alignment}\label{sec:case_study}
\begin{table}[!t]
\caption{When the excerpt from non-English (non-EN) Wiki pages is translated into EN, the answers are aligned more to the source context. With RAG, that is opposite.}
\centering
\small
\adjustbox{max width=\columnwidth}{
\begin{tabular}{cccccccc} 
\toprule
\multirow{2}{*}{\textbf{Target}} & \multirow{2}{*}{\textbf{Method}} & \multicolumn{2}{c}{\textbf{Llama}} & \multicolumn{2}{c}{\textbf{Aya}} & \multicolumn{2}{c}{\textbf{Qwen}} \\
\cmidrule(r){3-8}
 &  & non-EN & EN &  non-EN & EN  & non-EN & EN  \\
\midrule
\multirow{2}{*}{TR} & Base    & 17.28 & 22.46 & 15.34 & 20.91 &14.24 & 18.88 \\
                         & Wiki & 78.67 & 59.21 & 41.52 & 35.60 & 78.53 & 58.48 \\
                         & RAG & 23.51 & 33.71 & 15.55 & 21.45 & 15.91 & 27.10 \\
                         \midrule
\multirow{2}{*}{DE}  & Base  & 16.89 & 18.70 & 14.71 & 17.41 & 13.41 & 17.02  \\
                         & Wiki & 80.68 & 27.62 & 34.73 & 18.43 & 72.72 & 20.86 \\
                         & RAG & 23.75 & 34.69 & 15.89 & 19.84 & 15.65 & 30.21 \\
                        \midrule
\multirow{2}{*}{ZH} & Base & 22.44 & 28.13 & 16.89 & 23.87 & 15.47 & 21.74 \\
                         & Wiki & 83.14 & 51.39 & 43.67 & 34.40 & 77.99 & 46.95  \\
                         & RAG & 29.83 & 43.65 & 17.98 & 25.62 & 21.78 & 36.49  \\
                        \midrule
\multirow{2}{*}{IT} & Base  & 16.84 & 21.84 & 14.37 & 16.70 & 13.56 & 18.17 \\
                         & Wiki & 76.78 & 41.03 & 37.10 & 26.79 & 81.55 & 40.05 \\
                         & RAG & 24.56 & 40.05 & 15.31 & 21.06 & 15.83 & 20.29 \\
\bottomrule
\end{tabular}}
\label{fig:alignment_result}
\end{table}

Alignment of LLM responses with high-resource sources such as EN Wikipedia is often desirable, as health information in low-resource languages may be limited or lower quality~\cite{weissenberger2004breast,lawrentschuk37health,davaris2017thoracic}. In such cases, high-resource knowledge can help fill gaps. However, some scenarios require localized or domain-specific information, where EN-centric content may be unreliable or contextually inappropriate~\cite{jbp:/content/journals/10.1075/dt.25015.yan}. To explore this, we evaluated: $(i)$ providing contextual information directly into  the prompt, and $(ii)$ performing RAG~\cite{10.5555/3495724.3496517}. 
For the first method, we incorporated translated excerpts from non-EN Wikipedia pages that are semantically aligned with the given question. For RAG, we gathered scientific articles from PubMed,\footnote{\url{https://pubmed.ncbi.nlm.nih.gov/}} considered a reliable external source of health information, using Paperscraper,\footnote{\url{https://github.com/jannisborn/paperscraper}} querying by entity and nationality keywords (e.g., `allergy' + `Turkish') to obtain culturally specific information. We discarded entities with fewer than 50 retrieved PubMed articles and perform analysis on the rest. The RAG 
system, based on the simple HuggingFace implementation,
\footnote{\url{https://huggingface.co/blog/ngxson/make-your-own-rag}} employs the classical BM25 
model for sparse retrieval~\cite{bm25s}. We incorporated the top 10 articles that returned from the retriever into the prompt to enrich its context.

We compared the approaches against the baseline where the LLM responds to questions posed in EN (Section~\ref{exp:answer_analyze}). Both approaches improve reference alignment by producing more factual and concise answers than the baseline. As shown in Figure~\ref{fig:alignment_result}, LLMs incorporating excerpts from Wikipedia produce responses that align more closely with non-EN references, while with RAG, they align more with EN references. RAG results contain usually noisy context, leading to cases where LLMs are uncertain about their answers. Additionally, due to the increased prompt length, Aya was unable to generate responses for a few examples. Explicit, high-quality contextual information might be crucial for effective alignment. The setup represents an initial proof of concept rather than a comprehensive RAG evaluation, in future work, we plan to explore more advanced information retrieval and RAG methods to improve contextual relevance.
\section{Conclusion}
We introduced MultiWikiHealthCare, a multilingual benchmark for studying disparities across languages in healthcare Q\&A. By pairing popular queries with language-specific Wikipedia evidence, we quantified how differences in coverage (e.g., structure, citations, and fact availability) shape model behavior. The results show substantial variability across languages: baseline LLMs often privilege EN-centric evidence, while conditioning generation on source-language excerpts shifts grounding toward locally relevant knowledge. These findings highlight a practical path for improving equity in multilingual healthcare Q\&A, explicitly anchor answers in the user’s language of reference rather than defaulting to EN.
\\
\noindent
\textbf{Ethics Statement}
All analyses used publicly available Wikipedia data. Therefore, no private, sensitive, or personally identifiable health information was accessed or processed. All evaluations were performed solely for research purposes, and none of the LLMs analyzed should be solely used to provide a medical device. We acknowledge that the disparities observed across languages and Wikipedia may reflect broader inequities in global health communication and data representation. Through this analysis, we aim to promote more equitable and fair multilingual health-care applications.
\\
\noindent
\textbf{Limitations}
Finally, we acknowledge several limitations of our work: $(i)$ AlignScore is EN-centric, so we translate non-EN content into EN, which may introduce artifacts; $(ii)$ Wikipedia is the only reference source and not a clinical gold standard, so scores indicate alignment to Wikipedia rather than medical correctness or guideline adherence, without independent fact-checking; $(iii)$ budget constraints (paid APIs for most models) and limited native speakers restricted the number of Q\&A pairs and single-turn interactions; and $(iv)$ because LLMs are trained on heterogeneous sources, their knowledge may not align with Wikipedia, so our findings reflect alignment \emph{relative to Wikipedia} rather than clinical correctness. 
\\
\noindent
\textbf{Future Work}
\changemarker{We plan to extend reference sources beyond Wikipedia to include multilingual clinical and public health materials. Incorporating such sources would enable evaluation of factual alignment not only to publicly accessible knowledge but also to clinically validated recommendations. We will replace translation dependent scoring with multilingual factuality assessments complemented by native-speaker adjudication; and expand the scope to larger datasets, more languages, and multi-turn interactions.}

\begin{credits}
\subsubsection{\ackname}
\changemarker{The work of IBS, FMB and LF was supported by Lamarr Institute for Machine Learning and Artificial Intelligence. The work of PR was supported by the MARTINI project on Malicious Actors Profiling and Detection in Online Social Networks Through Artificial Intelligence funded by MCIN/AEI/ 10.13039/501100011033 and by EU NextGenerationEU/
PRTR. The work of BS and CB was supported by TANGO (Grant Agreement no. 101120763). Views and opinions expressed are, however, those of the author(s) only and do not necessarily reflect those of the European Union or the European Health and Digital Executive Agency (HaDEA). Neither the European Union nor the granting authority can be held responsible for them.}
\end{credits}

\raggedbottom
\bibliographystyle{splncs04}
\bibliography{ref.bib}

@article{yagnik2024medlm,
  title={MedLM: Exploring Language Models for Medical Question Answering Systems},
  author={Yagnik, Niraj and Jhaveri, Jay and Sharma, Vivek and Pila, Gabriel},
  journal={arXiv:2401.11389},
  year={2024}
}

@article{kim2025medical,
  title={Medical Hallucinations in Foundation Models and Their Impact on Healthcare},
  author={Kim, Yubin and Jeong, Hyewon and Chen, Shan and Li, Shuyue Stella and Park, Chanwoo and Lu, Mingyu and Alhamoud, Kumail and Mun, Jimin and Grau, Cristina and Jung, Minseok and others},
  journal={arXiv:2503.05777},
  year={2025}
}

@inproceedings{yu2024enhancing,
  title={{E}nhancing {H}ealthcare through {L}arge {L}anguage {M}odels: A {S}tudy on {M}edical {Q}uestion {A}nswering},
  author={Yu, Haoran and Yu, Chang and Wang, Zihan and Zou, Dongxian and Qin, Hao},
  booktitle={ICPICS},
  pages={895--900},
  year={2024},
  organization={IEEE}
}

@inproceedings{kaur-etal-2024-evaluating,
    title = "Evaluating Large Language Models for Health-related Queries with Presuppositions",
    author = "Kaur, Navreet  and
      Choudhury, Monojit  and
      Pruthi, Danish",
    booktitle = "Findings of the ACL",
    year = "2024",
    address = "Bangkok, Thailand",
    publisher = "ACL",
    pages = "14308--14331",
}

@inproceedings{koopman-zuccon-2023-dr,
    title = "Dr {C}hat{GPT} tell me what {I} want to hear: How different prompts impact health answer correctness",
    author = "Koopman, Bevan  and
      Zuccon, Guido",
    booktitle = "EMNLP",
    year = "2023",
        publisher = "ACL",
    address = "Singapore",
    pages = "15012--15022",
}

@inproceedings{sayin-etal-2024-llms,
    title = "Can {LLM}s Correct Physicians, Yet? Investigating Effective Interaction Methods in the Medical Domain",
    author = "Sayin, Burcu  and
      Minervini, Pasquale  and
      Staiano, Jacopo  and
      Passerini, Andrea",
    booktitle = "The 6th Clinical NLP",
    year = "2024",
    address = "Mexico City, Mexico",
    publisher = "ACL",
    pages = "218--237",
}

@article{weissenberger2004breast,
  title={{B}reast cancer: patient information needs reflected in {E}nglish and {G}erman web sites},
  author={Weissenberger, Christian and Jonassen, S and Beranek-Chiu, J and Neumann, M and M{\"u}ller, D and Bartelt, S and Schulz, S and M{\"o}nting, JS and Henne, K and Gitsch, G and others},
  journal={British J. of Cancer},
  volume={91},
  number={8},
  pages={1482--1487},
  year={2004},
  publisher={Nature Publishing Group}
}

@article{lawrentschuk37health,
  title={Health information quality on the internet in gynecological oncology: a multilingual evaluation},
  author={Lawrentschuk, N and Abouassaly, R and Hewitt, E and Mulcahy, A and Bolton, DM and Jobling, T},
  journal={Eur. J. Gynaecol. Oncol},
  volume={37},
  number={4},
  pages={2016}
}

@article{davaris2017thoracic,
  title={Thoracic surgery information on the internet: a multilingual quality assessment},
  author={Davaris, Myles and Barnett, Stephen and Abouassaly, Robert and Lawrentschuk, Nathan and others},
  journal={Interact J. Med Res},
  volume={6},
  number={1},
  pages={e6732},
  year={2017},
  publisher={JMIR Publications Inc., Toronto, Canada}
}

@inproceedings{jin2024better,
author = {Jin, Yiqiao and Chandra, Mohit and Verma, Gaurav and Hu, Yibo and De Choudhury, Munmun and Kumar, Srijan},
title = {Better to Ask in English: Cross-Lingual Evaluation of Large Language Models for Healthcare Queries},
year = {2024},
isbn = {9798400701719},
publisher = {ACM},
address = {New York, NY, USA},
booktitle = {The Web Conference},
pages = {2627–2638},
numpages = {12},
location = {Singapore, Singapore},
}

@inproceedings{xu-etal-2022-answer,
    title = "How Do We Answer Complex Questions: Discourse Structure of Long-form Answers",
    author = "Xu, Fangyuan  and
      Li, Junyi Jessy  and
      Choi, Eunsol",
    booktitle = "ACL (Volume 1: Long Papers)",
    year = "2022",
    address = "Dublin, Ireland",
    pages = "3556--3572",
}

@article{achiam2023gpt,
  title={Gpt-4 technical report},
  author={Achiam, Josh and Adler, Steven and Agarwal, Sandhini and Ahmad, Lama and Akkaya, Ilge and Aleman, Florencia Leoni and Almeida, Diogo and Altenschmidt, Janko and Altman, Sam and Anadkat, Shyamal and others},
  journal={arXiv:2303.08774},
  year={2023}
}

@article{dubey2024llama,
  title={The Llama 3 Herd of Models},
  author={Dubey, Abhimanyu and Jauhri, Abhinav and Pandey, Abhinav and Kadian, Abhishek and Al-Dahle, Ahmad and Letman, Aiesha and Mathur, Akhil and Schelten, Alan and Yang, Amy and Fan, Angela and others},
  journal={arXiv:2407.21783},
  year={2024}
}

@inproceedings{wolf2020transformers,
  author       = {Wolf, Thomas and Debut, Lysandre and Sanh, Victor and Chaumond, Julien and Delangue, Clement and Moi, Anthony and Cistac, Pierric and Rault, Tim and Louf, R{\'e}mi and Funtowicz, Morgan and others},
  title        = {Transformers: State-of-the-Art Natural Language Processing},
  booktitle    = {{EMNLP} (Demos)},
  pages        = {38--45},
  publisher    = {ACL},
  year         = {2020}
}

@inproceedings{DBLP:conf/www/ZhuA0R19,
author = {Zhu, Ming and Ahuja, Aman and Wei, Wei and Reddy, Chandan K.},
title = {A Hierarchical Attention Retrieval Model for Healthcare Question Answering},
year = {2019},
isbn = {9781450366748},
publisher = {ACM},
booktitle = {WWW},
pages = {2472–2482},
numpages = {11},
}

@inproceedings{DBLP:conf/trec/AbachaAPD17,
  author       = {Asma Ben Abacha and
                  Eugene Agichtein and
                  Yuval Pinter and
                  Dina Demner{-}Fushman},
  title        = {Overview of the Medical Question Answering Task at {TREC} 2017 LiveQA},
  booktitle    = {{TREC}},
  series       = {{NIST} Special Publication},
  volume       = {500-324},
  publisher    = {NIST},
  year         = {2017}
}

@inproceedings{DBLP:conf/medinfo/AbachaMSGSD19,
  author       = {Asma Ben Abacha and
                  Yassine Mrabet and
                  Mark Sharp and
                  Travis R. Goodwin and
                  Sonya E. Shooshan and
                  Dina Demner{-}Fushman},
  title        = {Bridging the Gap Between Consumers' Medication Questions and Trusted
                  Answers},
  booktitle    = {MedInfo},
  series       = {Studies in Health Technology and Informatics},
  volume       = {264},
  pages        = {25--29},
  publisher    = {{IOS} Press},
  year         = {2019}
}

@article{DBLP:journals/corr/abs-2309-01219,
  title={Siren’s Song in the AI Ocean: A Survey on Hallucination in Large Language Models},
  author={Zhang, Yue and Li, Yafu and Cui, Leyang and Cai, Deng and Liu, Lemao and Fu, Tingchen and Huang, Xinting and Zhao, Enbo and Zhang, Yu and Chen, Yulong and others},
  journal={Computational Linguistics},
  pages={1--46},
  year={2025},
  publisher={MIT Press 255 Main Street, 9th Floor, Cambridge, Massachusetts 02142, USA~…}
}

@inproceedings{schlicht2025llms,
author="Schlicht, Ipek Baris
and Zhao, Zhixue
and Sayin, Burcu
and Flek, Lucie
and Rosso, Paolo",
title="Do LLMs Provide Consistent Answers to Health-Related Questions Across Languages?",
booktitle="ECIR",
year="2025",
publisher="Springer Nature Switzerland",
address="Cham",
pages="314--322",
isbn="978-3-031-88714-7"
}

@inproceedings{samir-etal-2024-locating,
    title = "Locating Information Gaps and Narrative Inconsistencies Across Languages: A Case Study of {LGBT} People Portrayals on {W}ikipedia",
    author = "Samir, Farhan  and
      Park, Chan Young  and
      Field, Anjalie  and
      Shwartz, Vered  and
      Tsvetkov, Yulia",
    booktitle = "EMNLP 2024",
    month = nov,
    year = "2024",
    address = "Miami, Florida, USA",
    publisher = "ACL",
    pages = "6747--6762",
}

@inproceedings{park2021multilingual,
  title={Multilingual {C}ontextual {A}ffective {A}nalysis of {LGBT} {P}eople {P}ortrayals in {W}ikipedia},
  author={Park, Chan Young and Yan, Xinru and Field, Anjalie and Tsvetkov, Yulia},
  booktitle={ICWSM},
  volume={15},
  pages={479--490},
  year={2021}
}

@article{grattafiori2024llama,
  title={The {L}lama 3 {H}erd of {M}odels},
  author={Grattafiori, Aaron and Dubey, Abhimanyu and Jauhri, Abhinav and Pandey, Abhinav and Kadian, Abhishek and Al-Dahle, Ahmad and Letman, Aiesha and Mathur, Akhil and Schelten, Alan and Vaughan, Alex and others},
  journal={arXiv:2407.21783},
  year={2024}
}

@article{singhal_large_2023,
  title={Large language models encode clinical knowledge},
  author={Singhal, Karan and Azizi, Shekoofeh and Tu, Tao and Mahdavi, S Sara and Wei, Jason and Chung, Hyung Won and Scales, Nathan and Tanwani, Ajay and Cole-Lewis, Heather and Pfohl, Stephen and others},
  journal={Nature},
  volume={620},
  number={7972},
  pages={172--180},
  year={2023},
  publisher={Nature Publishing Group UK London}
}

@inproceedings{kwon2023efficient,
  title={Efficient Memory Management for Large Language Model Serving with PagedAttention},
  author={Woosuk Kwon and Zhuohan Li and Siyuan Zhuang and Ying Sheng and Lianmin Zheng and Cody Hao Yu and Joseph E. Gonzalez and Hao Zhang and Ion Stoica},
  booktitle={ACM SIGOPS},
  year={2023}
}

@inproceedings{restrepo2025multi,
  title={Multi-OphthaLingua: a multilingual benchmark for assessing and debiasing LLM ophthalmological QA in LMICs},
  author={Restrepo, David and Wu, Chenwei and Tang, Zhengxu and Shuai, Zitao and Phan, Thao Nguyen Minh and Ding, Jun-En and Dao, Cong-Tinh and Gallifant, Jack and Dychiao, Robyn Gayle and Artiaga, Jose Carlo and others},
  booktitle={AAAI},
  volume={39},
  number={27},
  pages={28321--28330},
  year={2025}
}

@article{gupta2025found,
  title={Found in Translation: Measuring Multilingual LLM Consistency as Simple as Translate then Evaluate},
  author={Gupta, Ashim and Mehta, Maitrey and Xu, Zhichao and Srikumar, Vivek},
  journal={arXiv:2505.21999},
  year={2025}
}

@article{schlicht_automatic_2024,
	title = {Automatic detection of health misinformation: a systematic review},
	volume = {15},
	issn = {1868-5137, 1868-5145},
	shorttitle = {Automatic detection of health misinformation},
	number = {3},
	urldate = {2025-06-04},
	journal = {J. AIHC},
	author = {Schlicht, Ipek Baris and Fernandez, Eugenia and Chulvi, Berta and Rosso, Paolo},
	month = mar,
	year = {2024},
	pages = {2009--2021},
}

@article{wang2025wikigap,
  title={WikiGap: Promoting Epistemic Equity by Surfacing Knowledge Gaps Between English Wikipedia and other Language Editions},
  author={Wang, Zining and Zhang, Yuxuan and Yoon, Dongwook and Vincent, Nicholas and Samir, Farhan and Shwartz, Vered},
  journal={arXiv:2505.24195},
  year={2025}
}

@inproceedings{10.5555/3495724.3496517,
author = {Lewis, Patrick and Perez, Ethan and Piktus, Aleksandra and Petroni, Fabio and Karpukhin, Vladimir and Goyal, Naman and K\"{u}ttler, Heinrich and Lewis, Mike and Yih, Wen-tau and Rockt\"{a}schel, Tim and Riedel, Sebastian and Kiela, Douwe},
title = {Retrieval-augmented generation for knowledge-intensive NLP tasks},
year = {2020},
isbn = {9781713829546},
publisher = {Curran Associates Inc.},
address = {Red Hook, NY, USA},
booktitle = {NIPS},
articleno = {793},
numpages = {16},
location = {Vancouver, BC, Canada},
series = {NIPS '20}
}

@misc{gpt5_2025,
  author       = {OpenAI},
  title        = {Introducing GPT-5},
  year         = {2025},
  month        = aug,
  howpublished = {\url{https://openai.com/index/introducing-gpt-5/}},
  note         = {Accessed 2025-09-29}
}

@article{gu2025surveyllmasajudge,
  title={A survey on llm-as-a-judge},
  author={Gu, Jiawei and Jiang, Xuhui and Shi, Zhichao and Tan, Hexiang and Zhai, Xuehao and Xu, Chengjin and Li, Wei and Shen, Yinghan and Ma, Shengjie and Liu, Honghao and others},
  journal={The Innovation},
year = {2026},
issn = {2666-6758},
doi = {https://doi.org/10.1016/j.xinn.2025.101253},
  publisher={Elsevier}
}

@inproceedings{zha-etal-2023-alignscore,
    title = "{A}lign{S}core: Evaluating Factual Consistency with A Unified Alignment Function",
    author = "Zha, Yuheng  and
      Yang, Yichi  and
      Li, Ruichen  and
      Hu, Zhiting",
    booktitle = "ACL",
    month = jul,
    year = "2023",
    address = "Toronto, Canada",
    publisher = "ACL",
    pages = "11328--11348", 
}

@article{jbp:/content/journals/10.1075/dt.25015.yan,
   author = "Yang, Tian and Valdez, Susana",
   title = "{H}ow machine translation is used in healthcare",
   journal = "Digital Translation",
   issn = "2949-6861",
   year = "2025",
   publisher = "John Benjamins",
   doi = "https://doi.org/10.1075/dt.25015.yan",
}

@article{liu2019roberta,
  title={Roberta: A robustly optimized bert pretraining approach},
  author={Liu, Yinhan and Ott, Myle and Goyal, Naman and Du, Jingfei and Joshi, Mandar and Chen, Danqi and Levy, Omer and Lewis, Mike and Zettlemoyer, Luke and Stoyanov, Veselin},
  journal={arXiv:1907.11692},
  year={2019}
}

@article{ford2004content,
  title={Content analysis: An introduction to its methodology},
  author={Ford, John M},
  journal={Personnel psychology},
  volume={57},
  number={4},
  pages={1110},
  year={2004},
  publisher={Blackwell Publishing Ltd.}
}

@article{guo2025deepseek,
  title={DeepSeek-R1 incentivizes reasoning in LLMs through reinforcement learning},
  author={Guo, Daya and Yang, Dejian and Zhang, Haowei and Song, Junxiao and Wang, Peiyi and Zhu, Qihao and Xu, Runxin and Zhang, Ruoyu and Ma, Shirong and Bi, Xiao and others},
  journal={Nature},
  volume={645},
  number={8081},
  pages={633--638},
  year={2025},
  publisher={Nature Publishing Group UK London}
}

@article{dang2024aya,
  title={Aya expanse: Combining research breakthroughs for a new multilingual frontier},
  author={Dang, John and Singh, Shivalika and D'souza, Daniel and Ahmadian, Arash and Salamanca, Alejandro and Smith, Madeline and Peppin, Aidan and Hong, Sungjin and Govindassamy, Manoj and Zhao, Terrence and others},
  journal={arXiv:2412.04261},
  year={2024}
}

@inproceedings{devlin-etal-2019-bert,
    title = "{BERT}: Pre-training of Deep Bidirectional Transformers for Language Understanding",
    author = "Devlin, Jacob  and
      Chang, Ming-Wei  and
      Lee, Kenton  and
      Toutanova, Kristina",
    booktitle = "ACL",
    month = jun,
    year = "2019",
    address = "Minneapolis, Minnesota",
    publisher = "ACL",
    pages = "4171--4186",
}

@article{tierney2025health,
  title={{H}ealth {E}quity in the {E}ra of {L}arge {L}anguage? {M}odels.},
  author={Tierney, Aaron A and Reed, Mary E and Grant, Richard W and Doo, Florence X and Pay{\'a}n, Denise D and Liu, Vincent X},
  journal={American Journal of Managed Care},
  volume={31},
  number={3},
  year={2025}
}

@article{nigatu2025into,
  title={{I}nto the {V}oid: {U}nderstanding {O}nline {H}ealth {I}nformation in {L}ow-{W}eb {D}ata {L}anguages},
  author={Nigatu, Hellina Hailu and Abdelkadir, Nuredin Ali and Tewelde, Fiker and Chancellor, Stevie and Wilkinson, Daricia},
  journal={arXiv:2509.20245},
  year={2025}
}

@article{hurst2024gpt,
  title={{GPT}-4o {S}ystem {C}ard},
  author={Hurst, Aaron and Lerer, Adam and Goucher, Adam P and Perelman, Adam and Ramesh, Aditya and Clark, Aidan and Ostrow, AJ and Welihinda, Akila and Hayes, Alan and Radford, Alec and others},
  journal={arXiv:2410.21276},
  year={2024}
}

@article{bm25s,
  title={Bm25s: Orders of magnitude faster lexical search via eager sparse scoring},
  author={L{\`u}, Xing Han},
  journal={arXiv:2407.03618},
  year={2024}
}

@article{yang2025qwen3,
  title={Qwen3 technical report},
  author={Yang, An and Li, Anfeng and Yang, Baosong and Zhang, Beichen and Hui, Binyuan and Zheng, Bo and Yu, Bowen and Gao, Chang and Huang, Chengen and Lv, Chenxu and others},
  journal={arXiv:2505.09388},
  year={2025}
}

@inproceedings{es-etal-2024-ragas,
    title = "{RAGA}s: Automated Evaluation of Retrieval Augmented Generation",
    author = "Es, Shahul  and
      James, Jithin  and
      Espinosa Anke, Luis  and
      Schockaert, Steven",
    booktitle = "EACL: System Demonstrations",
    month = mar,
    year = "2024",
    address = "St. Julians, Malta",
    doi = "10.18653/v1/2024.eacl-demo.16",
    pages = "150--158",
}

@inproceedings{ranathunga-de-silva-2022-languages,
    title = "Some Languages are More Equal than Others: Probing Deeper into the Linguistic Disparity in the {NLP} World",
    author = "Ranathunga, Surangika  and
      de Silva, Nisansa",
    booktitle = "AACL-IJCNLP (Volume 1: Long Papers)",
    month = nov,
    year = "2022",
    address = "Online",
    publisher = "ACL",
    pages = "823--848",
}

@article{wissler1905spearman,
  title={The Spearman correlation formula},
  author={Wissler, Clark},
  journal={Science},
  volume={22},
  number={558},
  pages={309--311},
  year={1905},
  publisher={American Association for the Advancement of Science}
}

@incollection{ross2017paired,
  title={Paired samples T-test},
  author={Ross, Amanda and Willson, Victor L},
  booktitle={Basic and advanced statistical tests: Writing results sections and creating tables and figures},
  pages={17--19},
  year={2017},
  publisher={Springer}
}

@article{agarwal2024medhalu,
  title={MedHalu: Hallucinations in Responses to Healthcare Queries by Large Language Models},
  author={Agarwal, Vibhor and Jin, Yiqiao and Chandra, Mohit and De Choudhury, Munmun and Kumar, Srijan and Sastry, Nishanth},
  journal={arXiv:2409.19492},
  year={2024}
}

@inproceedings{xlmroberta,
    title = "Unsupervised Cross-lingual Representation Learning at Scale",
    author = "Conneau, Alexis  and
      Khandelwal, Kartikay  and
      Goyal, Naman  and
      Chaudhary, Vishrav  and
      Wenzek, Guillaume  and
      Guzm{\'a}n, Francisco  and
      Grave, Edouard  and
      Ott, Myle  and
      Zettlemoyer, Luke  and
      Stoyanov, Veselin",
    editor = "Jurafsky, Dan  and
      Chai, Joyce  and
      Schluter, Natalie  and
      Tetreault, Joel",
    booktitle = "Proceedings of the 58th Annual Meeting of the Association for Computational Linguistics",
    month = jul,
    year = "2020",
    address = "Online",
    publisher = "Association for Computational Linguistics",
    url = "https://aclanthology.org/2020.acl-main.747/",
    doi = "10.18653/v1/2020.acl-main.747",
    pages = "8440--8451"
}

\end{document}